\documentclass{article}

\usepackage[main, final]{neurips_2026}
\usepackage[utf8]{inputenc} 
\usepackage[T1]{fontenc}    
\usepackage{hyperref}       
\usepackage{url}            
\usepackage{booktabs}       
\usepackage{amsfonts}       
\usepackage{nicefrac}       
\usepackage{microtype}      
\usepackage{xcolor}         

\usepackage{graphicx} 
\usepackage{amsmath}
\usepackage{multirow}
\usepackage[table]{xcolor}
\usepackage[titles]{tocloft}
\usepackage[titletoc, title]{appendix}
\usepackage{titletoc}
\usepackage{enumitem}
\usepackage{wrapfig}
\usepackage{booktabs}

\title{Beyond Text Conditioning: A Systematic Study of MLLM-DiT Fusion for Video Generation}

\author{%
    Yanbo Ding\textsuperscript{\rm1,2}\thanks{These authors contributed equally, and this work was done during internship at Microsoft Research.}
    \ , \ 
    Yijia Fan\textsuperscript{\rm2,3$*$}, \ 
    Caihua Shan\textsuperscript{\rm2}\thanks{Corresponding authors.} 
    \ , \ 
    Yifan Yang\textsuperscript{\rm2} 
    , \ 
    Yifei Shen\textsuperscript{\rm2} 
    , \\
    \textbf{Weijie Wang}\textsuperscript{\rm4} 
    , \ 
    \textbf{Xirui Hu}\textsuperscript{\rm5} 
    , \ 
    \textbf{Dongsheng Li}\textsuperscript{\rm2} 
    , \ 
    \textbf{Lili Qiu}\textsuperscript{\rm2} 
    , \ 
    \textbf{Yuqing Yang}\textsuperscript{\rm2} 
    , \ 
    \textbf{Yali Wang}\textsuperscript{\rm1,6$\dag$} \\
    \\
    \textsuperscript{\rm 1}Chinese Academy of Sciences \\
    \textsuperscript{\rm 2}Microsoft Research \\
    \textsuperscript{\rm 3}Sun Yat-sen University \\
    \textsuperscript{\rm 4}Zhejiang University \\
    \textsuperscript{\rm 5}Xi’an Jiaotong University \\
    \textsuperscript{\rm 6}Shanghai Artificial Intelligence Laboratory \\
}

\begin{document}

\maketitle



\begin{abstract}

Diffusion Transformers (DiTs) have become the dominant paradigm for high-fidelity video generation, yet their ability to perform high-level semantic planning remains limited. While hybrid architectures integrating MLLMs with diffusion backbones have shown strong advantages in image synthesis, such designs remain underexplored in video generation, where existing approaches often treat MLLMs primarily as frozen feature encoders rather than semantic generators. To fill this gap, we systematically study how an MLLM should be integrated with a DiT for video generation by answering three questions: what intermediate representation should bridge the MLLM and DiT, how the MLLM should generate it, and how the DiT should incorporate it during diffusion rendering. Our analysis reveals three key findings: (1) discrete semantic visual tokens produced by an EMA-based tokenizer provide a stable and expressive interface, (2) autoregressive causal modeling is effective for generating these tokens, and (3) explicit visual-token conditioning is more effective than prompt refinement or latent bridging. Based on these findings, we propose BiVidGen, a hybrid framework where an MLLM first generates semantic visual tokens and a DiT renders videos conditioned on both text and these tokens via multi-layer cross-attention. Extensive experiments show that BiVidGen improves semantic alignment and temporal coherence over a fine-tuned DiT baseline, achieving stronger performance on VBench-Long. These results suggest that explicit MLLM-based visual planning provides an effective intermediate interface for text-to-video generation beyond text-only conditioning.



\end{abstract}

\section{Introduction}
\label{sec:introduction}

Video generative models have advanced rapidly in recent years. From early large-scale demonstrations such as Sora~\citep{openai2024sora} to recent open-sourced models like CogVideo~\citep{yang2024cogvideox,hong2022cogvideo}, HunyuanVideo~\citep{kong2024hunyuanvideo}, and Wan~\citep{wan2025wan}, diffusion-based transformers (DiTs)~\citep{peebles2023dit} have emerged as the dominant paradigm and established the state-of-the-art foundation for high-fidelity video synthesis.

In parallel, hybrid architectures that integrate multimodal large language models (MLLMs)~\citep{bai2025qwen3vltechnicalreport} with powerful diffusion backbones have reshaped the generation paradigm. Representative works for image generation include MetaQuery~\citep{pan2025metaquery}, Qwen-Image~\citep{wu2025qwen}, BLIP3o-Next~\citep{chen2025blip3o-next}, and X-Omni~\citep{geng2025xomni}, which demonstrate that combining language-centric models with high-capacity visual decoders significantly enhances text understanding and fine-grained text-visual alignment. Beyond conventional text conditioning, MLLMs are capable of providing richer and more structured intermediate representations for visual synthesis.


Despite their success in images, such hybrid designs remain largely underexplored in video generation. More recently, unified video models~\citep{wei2025univideo, luo2025univid, tan2025omnivideo, yang2026omni} have incorporated MLLMs, but typically freeze the language backbone and rely on lightweight adapters or learnable query tokens. As a result, these methods still treat the MLLM primarily as a feature encoder, rather than fully exploiting its generative capability for structured semantic planning.


This raises a central question: beyond conventional text conditioning, how should an MLLM be effectively integrated with a DiT for video generation? We systematically study this problem through three progressively dependent questions: (1) what form of information should be passed from the MLLM to the DiT; (2) how such information should be modeled and generated by the MLLM; and (3) how the DiT should leverage it for diffusion-based rendering. Together, these questions define the overall pathway from high-level semantic planning to low-level visual synthesis.

\textbf{Intermediate Representation Between MLLM and DiT.} 
Existing choices include natural-language descriptions, where the MLLM refines or expands the prompt, and latent tokens, where hidden states or learnable queries provide implicit conditioning signals. To better exploit the generative potential of MLLMs, we study explicit visual-semantic tokens planned by the MLLM as a more direct intermediate representation for DiT rendering. A natural option is to use the continuous visual features originally adopted by MLLMs, which preserve rich information. We further explore discrete semantic tokens, which provide a compact interface better aligned with autoregressive generation while remaining sufficient for video-level semantics. For discretization, we compare EMA-based~\citep{hunter1986ema} and embedding-based tokenizers, with or without auxiliary supervision such as MLLM understanding loss or pixel reconstruction loss. We find that the EMA-based tokenizer offers the most stable representation, preserving the original feature scale without gradient-based codebook updates or extra supervision. We also explore hierarchical token structures inspired by Qwen3-VL \citep{bai2025qwen3vltechnicalreport}, which encode multi-level visual information from shallow perceptual features to deep semantic representations, but they bring limited gains for generation in our experiments.

\textbf{MLLM Generation Mechanism.} We next examine how the MLLM should generate intermediate visual information. Specifically, we compare causal attention with full attention over visual tokens. Causal attention follows the autoregressive structure of language modeling and naturally aligns with next-token prediction, making it suitable for sequential video dynamics. Full attention, by contrast, enables bidirectional interactions among all visual tokens and may improve holistic representation learning. Our experiments show that causal attention is more effective for semantic token generation, producing more structured temporal representations while maintaining strong visual quality.

\textbf{MLLM-DiT Fusion Strategy.} Finally, we study how the DiT should leverage the information produced by the MLLM during diffusion rendering. We compare explicit visual token conditioning with latent bridging through learnable queries and a connector module. For explicit conditioning, the MLLM-generated tokens are injected into the DiT through either self-attention or cross-attention, allowing them to provide structural guidance throughout denoising. Our results show that explicit visual-token conditioning, especially with multi-layer cross-attention, offers stronger guidance for the DiT and achieves better generation performance than query-based latent bridging.

In summary, based on extensive and systematic experimental analysis, we propose BiVidGen, which adopts discrete semantic tokens produced by an EMA-based tokenizer, maintains causal attention within the MLLM, and uses explicitly generated visual tokens as structured conditions for the DiT. 
Quantitative results on VBench-Long~\citep{huang2024vbench} demonstrate its superior performance, while qualitative results further show that MLLM-based semantic planning helps the DiT render videos with improved temporal consistency, causal transitions, and semantic alignment.

\section{Related Work}
\label{sec:related_work}

\paragraph{Image Generation.}  

Diffusion-based image generation models have rapidly advanced, 
from the Stable Diffusion series~\citep{rombach2022sd,podell2023sdxl,esser2024sd3} to Flux~\citep{flux2024,labs2025flux1kontextflowmatching} and Z-Image~\citep{cai2025z-image}.
More recently, hybrid architectures that combine autoregressive models with diffusion decoders have emerged.  These methods can be grouped into two categories: integrated diffusion heads and external diffusion backbones. The first paradigm unifies text and image modeling. 
Methods such as Transfusion~\citep{zhou2024transfusion}, Orthus~\citep{kou2024orthus}, JanusFlow~\citep{ma2025janusflow}, Emu3.5~\citep{cui2025emu3}, Bagel~\citep{deng2025bagel} and Show-o2~\citep{xie2025showo2} adopt an autoregressive backbone for text modeling, and delegate image generation to a diffusion loss or a flow-based head~\citep{lipman2022flow} on continuous visual latents. 

Another line of work couples a pretrained MLLM with an external diffusion decoder. Representative examples include MetaQuery~\citep{pan2025metaquery}, Qwen-Image~\citep{wu2025qwen}, and Open-Uni~\citep{wu2025openuni}. Instead of integrating diffusion loss within a single model, these methods use the MLLMs to produce intermediate representations that are subsequently decoded by an independent diffusion model. While both keep the MLLM backbone frozen, MetaQuery~\citep{pan2025metaquery} and Open-Uni~\citep{wu2025openuni} train lightweight query tokens to extract conditions for the diffusion decoder, whereas Qwen-Image~\citep{wu2025qwen} directly uses the MLLM’s output text hidden states as diffusion conditions. In contrast, BLIP3o-Next~\citep{chen2025blip3o-next}, X-Omni~\citep{geng2025xomni}, and the latest work on multimodal pretraining~\citep{tong2026languagemodelingexplorationmultimodal} retrain the MLLM to explicitly generate semantically meaningful visual tokens, which provide stronger visual conditions. 

Our work is most closely related to the latter paradigm, retraining MLLMs for visual token generation. However, rather than directly adopting a single design choice, we systematically study a range of design variants, including integrated diffusion heads within MLLMs and frozen-MLLM designs.

\paragraph{Video Generation.} 

Video generation has developed along two main paradigms: diffusion-based models and autoregressive models. From early U-Net-based~\citep{ronneberger2015unet} methods like SVD~\citep{blattmann2023svd} to more recent transformer-based approaches such as Open-Sora~\citep{zheng2024opensora,lin2024opensora-plan,peng2025opensora2}, HunyuanVideo~\citep{kong2024hunyuanvideo}, and Wan~\citep{wan2025wan}, diffusion-based methods achieve strong visual fidelity. Meanwhile, autoregressive approaches, including Loong~\citep{wang2024loong}, MAGI-1~\citep{teng2025magi}, and VideoAR~\citep{ji2026videoar}, generate frames sequentially, enabling flexible length extension and temporal causality. However, these approaches are typically built upon visual tokens from (VQ)VAE-based tokenizers~\citep{VQVAE}, rather than leveraging semantically enriched visual tokens from the MLLM.

Recently, unified video models (e.g., UniVideo~\citep{wei2025univideo}, UniVid~\citep{luo2025univid}, Omni-Video~\citep{tan2025omnivideo, yang2026omni}) have incorporated MLLMs together with DiTs for video generation and editing. However, these approaches largely freeze the MLLM parameters to preserve their understanding capabilities, introducing only lightweight components such as learnable query tokens, additional vision heads or adapters. In contrast, we retrain the entire MLLM on visual generation data, enabling it to produce semantic-level visual tokens for DiT-based rendering.


\section{Method}
\label{sec:method}


\paragraph{Overview.}



To explore hybrid MLLM-DiT architectures, we first formulate the overall generation process. Given a text prompt $T$, a standard text-conditioned DiT directly models the generation process as $p(X \mid T)$, where $X$ denotes the output video. In contrast, we introduce an intermediate representation $D$ produced by the MLLM prior to diffusion rendering. Specifically, the MLLM generates $D \sim p_\phi(D \mid T)$ from the input prompt, and the DiT then models $X \sim p_\theta(X \mid D, T)$ conditioned on both the MLLM-produced representation and the original text prompt.


\subsection{The Form of Intermediate Representation}
\label{sec:tokenizer}

\begin{figure*}[!t]
  \begin{center}
    \centerline{\includegraphics[width=1.0\textwidth]{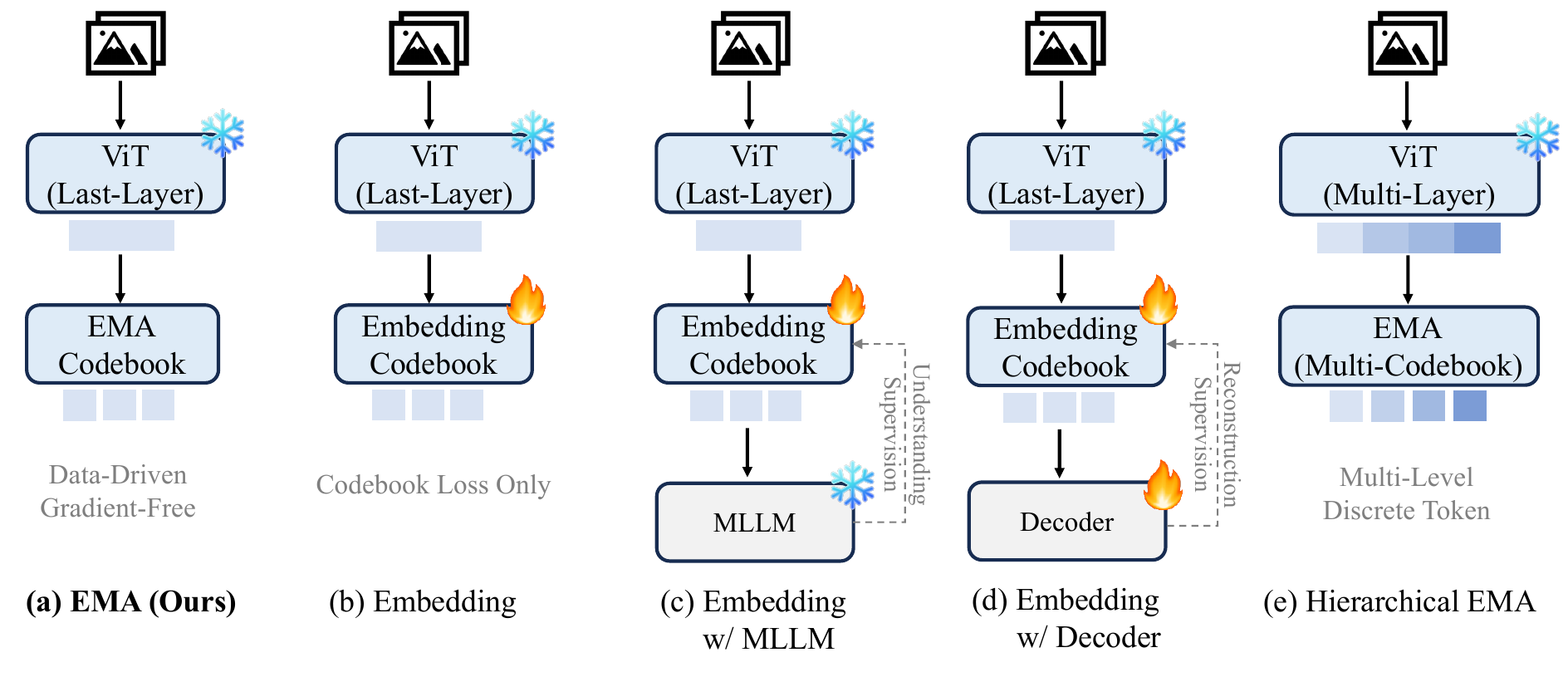}}
    \vspace{-8pt}
    \caption{
    \textbf{Quantization Design Variants.} Codebooks include (a) EMA-based updates, (b) gradient-based embedding updates, (c) embedding updates with MLLM understanding supervision, (d) embedding updates with pixel reconstruction supervision, and (e) hierarchical EMA-based updates.
    }
    \label{fig:vq-ablation}
  \end{center}
\end{figure*}

The intermediate representation $D$ that bridges the MLLM and the DiT can take several forms, including natural language, implicit latent tokens, and explicit visual tokens. The first two forms have been explored in prior work; here, we focus on the choice of explicit visual tokens.

Since the original MLLM is not designed to directly generate visual-semantic tokens, enabling explicit visual-token generation requires adapting its output space. Specifically, we first need to decide which visual space the MLLM should generate in, and then retrain the MLLM to autoregressively produce tokens in this space conditioned on text prompts. A natural starting point is to leverage the visual encoder inherited from the MLLM to extract continuous semantic visual features. These continuous features preserve rich visual details and are aligned with the MLLM's multimodal representation space. However, they are not naturally compatible with discrete next-token prediction, making them difficult to model directly with an autoregressive MLLM. We therefore study whether these continuous semantic features can be quantized into discrete visual tokens, similar to text tokens.

As illustrated in Figure~\ref{fig:vq-ablation}, we compare five quantization variants. Specifically, let $\mathbf{C}$ denote the continuous visual feature sequence extracted by the MLLM visual encoder and $\{\mathbf{e}_n\}_{n=1}^N$ denote a learned visual codebook. Each feature vector $\mathbf{C}_k$ is quantized to its nearest codebook entry as $\mathbf{D}_k = \mathbf{e}_{d_k}$, where $d_{k} = \arg\min_{n} \|\mathbf{C}_{k} - \mathbf{e}_n\|_2$ and $k$ indexes the flattened visual feature sequence.

For embedding-based tokenizers illustrated in Figure~\ref{fig:vq-ablation} (b)--(d), we optimize the standard vector-quantization objective $\mathcal{L}_{\mathrm{VQ}} = \sum_{k} \|\mathrm{sg}[\mathbf{C}_k] - \mathbf{D}_k\|_2^2 + \beta \|\mathbf{C}_k - \mathrm{sg}[\mathbf{D}_k]\|_2^2$, where $\mathrm{sg}[\cdot]$ denotes the stop-gradient operator and $\beta$ controls the commitment loss. We additionally consider two auxiliary supervision variants: an MLLM understanding loss $\mathcal{L}_{\mathrm{MLLM}} = -\sum_{j=1}^{M} \log p_{\psi}(y_j \mid y_{<j}, \mathbf{D})$, where a frozen MLLM $p_\psi$ predicts the paired text sequence $Y$ from the quantized features; and a reconstruction loss $\mathcal{L}_{\mathrm{rec}} = \|g_{\omega}(\mathbf{D}) - X\|_2$, where a decoder $g_\omega$ reconstructs the original video $X$.

Moreover, we consider an alternative updating scheme based on Exponential Moving Average (EMA)~\citep{hunter1986ema} without gradient-based optimization (Figure~\ref{fig:vq-ablation} (a)). It maintains the exponential moving averages $\mathbf{s}_n$ of the accumulated features and assignment counts $c_n$ for the $n$-th code:


\begin{equation} 
\mathbf{s}_n^{(i)} = \mu \mathbf{s}_n^{(i-1)} + (1-\mu) \sum_{k:d_k=n}\mathbf{C}_k, \quad
c_n^{(i)} = \mu c_n^{(i-1)} + (1-\mu) |\{k:d_k=n\}| 
\end{equation}
where $\mu$ is the decay factor, and $i$ is the current training step. The codebook entry is updated as
$
\mathbf{e}_n =
\frac{\mathbf{s}_n^{(i)}}{c_n^{(i)}+\delta},
$
where $\delta$ is a small constant for stability.
This EMA update rule makes the codebook statistics data-driven, which helps preserve the feature distribution and reduce quantization error.


Finally, we explore hierarchical token structures adopted in Qwen3-VL~\citep{bai2025qwen3vltechnicalreport}, where features from different layers of the visual encoder are quantized to capture different levels of semantics (Figure~\ref{fig:vq-ablation} (e)).
The details of hierarchical token structures are provided in Appendix~\ref{sec:hierarchical_token}.

\subsection{MLLM-based Generation Mechanisms}
\label{sec:mllm}

After deciding the intermediate representation, we next study how the MLLM should produce it from the input prompt. For natural-language representations, the MLLM follows its original language generation ability and outputs a refined or expanded prompt. For implicit latent-token interfaces, the MLLM is typically kept frozen, while learnable queries or adapters extract hidden conditioning features from its internal states. 
However, these approaches fall short of fully exploiting the generative capability of the MLLM for structured semantic planning.
We therefore focus on explicit visual-token generation, which requires the MLLM to acquire a new visual generation capability. To this end, we fine-tune the MLLM to explicitly generate semantic-visual tokens conditioned on the text prompt. This involves two key design questions: what attention mechanism should the MLLM use when generating visual tokens, and whether the generated tokens should be discrete or continuous.



\begin{figure*}[!t]
  \begin{center}
    \centerline{\includegraphics[width=1.0\textwidth]{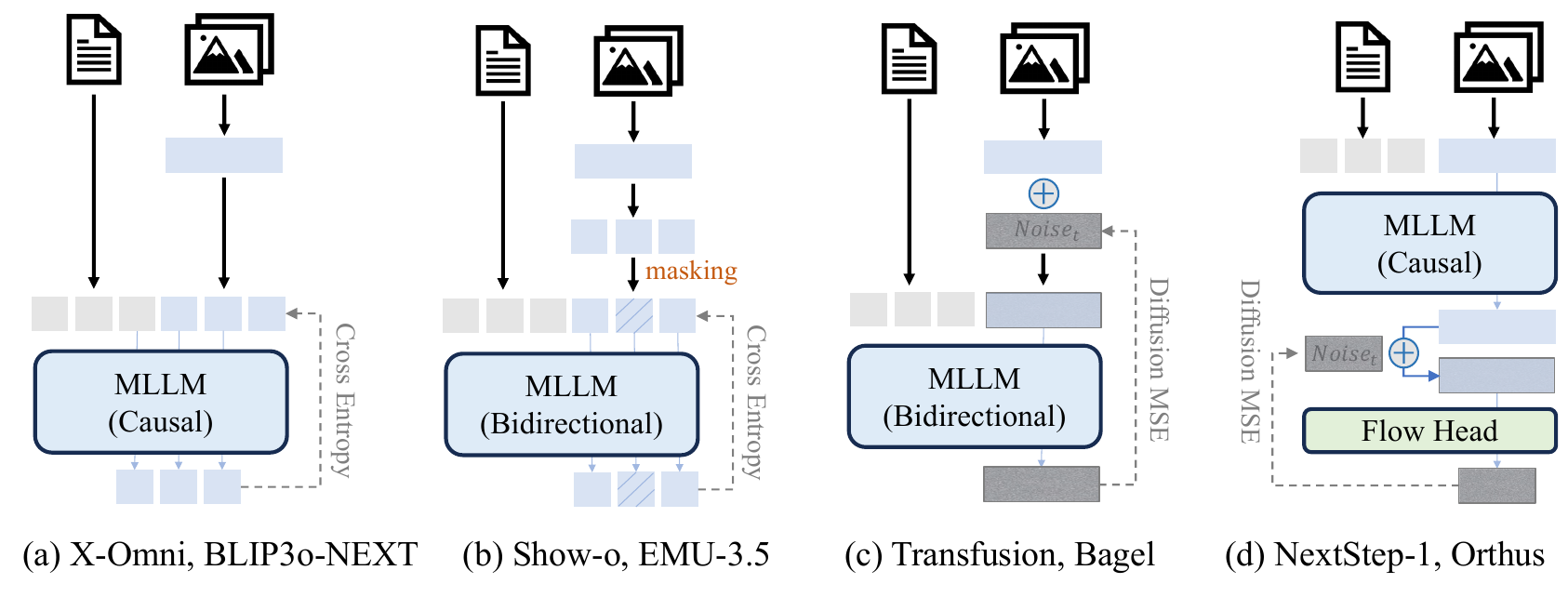}}
    \vspace{-8pt}
    \caption{
    \textbf{MLLM-Based Generation Variants.}
    We comprehensively and fairly compare mainstream modeling strategies for visual generation using MLLMs, including causal versus bidirectional attention mechanisms, and continuous versus discrete input representations. 
    }
    \label{fig:ar-ablation}
  \end{center}
\end{figure*}

In Figure~\ref{fig:ar-ablation}, we compare a set of multimodal architectures that unify discrete text tokens with discrete or continuous visual tokens, inspired by recent unified image models~\citep{deng2025bagel,geng2025xomni,cui2025emu3,team2025nextstep}.
We design four representative architectures: 
(a) maintain causal attention in the MLLM and treat both discrete text tokens and visual tokens as a unified autoregressive sequence; 
(b) apply bidirectional attention to discrete visual tokens, where the masked positions are randomly selected but kept consistent across frames, and the model predicts the masked tokens; 
(c) adopt bidirectional attention over continuous visual tokens corrupted with noise, where the model is trained to predict the added noise under a diffusion objective; 
(d) maintain causal attention for multimodal token processing and introduce a flow matching head~\citep{labs2025flux1kontextflowmatching} that uses the MLLM's visual outputs as conditions to generate the final continuous visual tokens.

\subsection{Connecting DiT-Based Renderer}
\label{sec:dit}

The third key question concerns how the intermediate tokens produced by the MLLM should be exposed to the DiT renderer. We consider three major categories of MLLM-DiT connection strategies: natural-language prompt refinement, implicit latent bridging, and explicit visual-token conditioning.

\begin{figure*}[!t]
  \begin{center}
    \centerline{\includegraphics[width=1.0\textwidth]{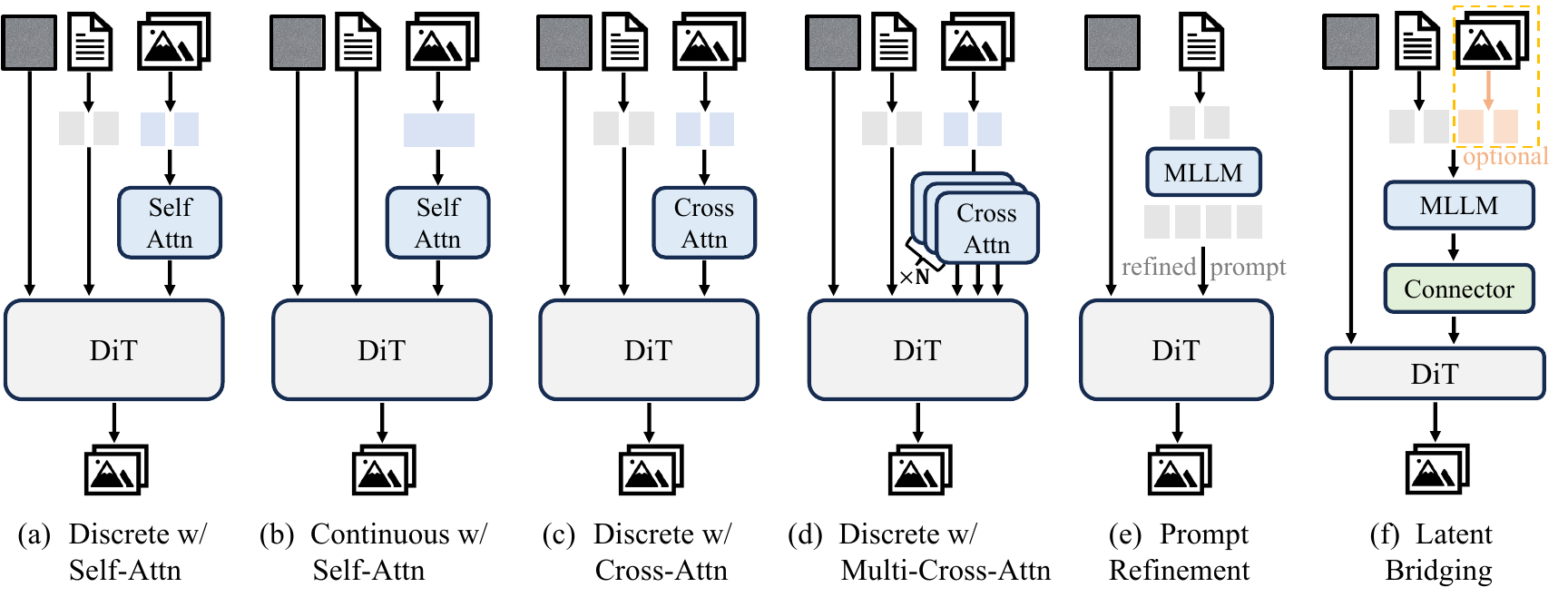}}
    \vspace{-7pt}
    \caption{
    \textbf{DiT-Based Generation Variants.} We categorize MLLM-DiT connection strategies into (a-d) explicit visual-token conditioning, including self-attention, cross-attention, and multi-layer injection;
(e) prompt refinement, which transfers MLLM information through text only; and
(f) latent bridging, which uses learnable queries and a connector to extract implicit MLLM tokens.}
    \label{fig:connect-ablation}
  \end{center}
\end{figure*}

For natural-language representations (Figure~\ref{fig:connect-ablation} (e)), the MLLM rewrites or expands the original prompt, and the standard DiT generates videos using the refined text prompt. This strategy is simple and straightforward but the transferred information remains limited to the language space, and the refined prompt does not provide any fine-grained visual information to the DiT-based renderer.

For implicit latent bridging (Figure~\ref{fig:connect-ablation} (f)), learnable query tokens extract hidden states from the MLLM, and a connector maps the queried features to DiT-compatible conditions. This design follows the common frozen-MLLM paradigm, which may lead to a narrow information bottleneck.


For explicit visual-token conditioning (Figure~\ref{fig:connect-ablation} (a)--(d)), we investigate how MLLM-generated visual tokens can be effectively injected into the DiT from three perspectives: token format, injection mechanism, and injection depth. First, we compare discrete and continuous visual tokens to examine how token representations affect generation and reconstruction quality. Second, we study two injection mechanisms: self-attention and cross-attention. When incorporated via self-attention, the generated visual tokens act as structural blueprints and directly participate in the denoising dynamics. In contrast, when introduced via cross-attention, they serve as conditioning signals that guide the denoising process externally. Finally, to encourage more effective utilization of visual conditions, we explore multi-layer injection, where visual guidance is introduced into multiple DiT blocks.

\begin{figure*}[!t]
  \begin{center}
    \centerline{\includegraphics[width=1.0\textwidth]{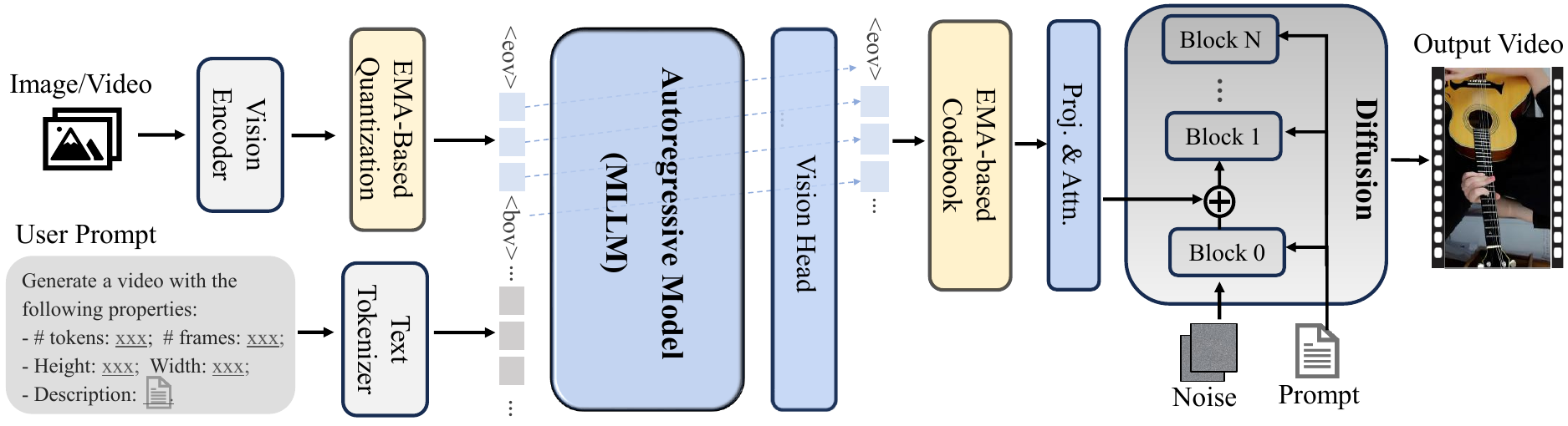}}
    \vspace{-5pt}
    \caption{
    \textbf{Overview.} BiVidGen consists of three main components. First, an EMA-based vision tokenizer provides discrete visual representations. Then, an MLLM predicts semantic visual tokens. Finally, the DiT renders high-fidelity videos conditioned on these planned tokens.
    }
    \label{fig:overview}
  \end{center}
\end{figure*}

\subsection{Final Architectures}

After systematically analyzing the key design choices, we now present the final architecture. As illustrated in Figure~\ref{fig:overview}, BiVidGen consists of an EMA-based tokenizer, an MLLM-based semantic generator, and a DiT-based renderer, trained separately and connected sequentially during inference. Since the EMA-based tokenizer has been described in Section~\ref{sec:tokenizer}, we omit its details here and focus on the training procedures for the MLLM and the DiT, as well as the overall inference process.

\textbf{MLLM Training.} We fine-tune the MLLM on paired text--image or text--video data. A resolution--frame prefix is prepended to the visual token sequence to improve control over resolution and frame count (Figure~\ref{fig:overview}). The visual tokens are obtained from our EMA-based tokenizer, appended after the text tokens and enclosed by \texttt{<bov>} and \texttt{<eov>} markers. We optimize the model using standard next-token prediction with teacher forcing~\citep{vaswani2017attention}, where the cross-entropy loss is applied only to the visual tokens and the \texttt{<eov>} token, ensuring correct sequence termination.


\textbf{DiT Training.} The DiT is conditioned on discrete visual tokens $\mathbf{D}$ from the EMA-based tokenizer. $\mathbf{D}$ is projected to the DiT hidden dimension and injected into each DiT block via cross-attention:
\begin{equation}
\mathbf{X}_t^{b+1} = \mathbf{X}_t^b + \text{CrossAttn}(\mathbf{X}_t^b, \text{Proj}(\mathbf{D})),
\end{equation}
where $b$ indexes the DiT blocks. By introducing visual guidance at every block, the DiT integrates semantic conditions throughout the entire denoising process, enabling stronger and more fine-grained control. To preserve the pretrained diffusion prior, we freeze the DiT backbone and train only the newly added cross-attention modules and the projection layer with the standard diffusion MSE loss.


\textbf{Inference Process.}
During inference, the MLLM autoregressively generates visual tokens starting from the \texttt{<bov>} marker, conditioned on the user prompt. The resulting visual token sequence is then fed into the diffusion generator to produce the final visual content. To ensure stable spatio-temporal modeling, we retain the original 3D RoPE~\citep{su2024roformer} positional encoding for visual tokens.

\section{Experiment}
\label{sec:experiment}

\subsection{Experimental Setup}
\label{sec:setup}

\paragraph{Training Details.} 
We use Qwen3-VL-2B~\citep{bai2025qwen3vltechnicalreport} as the base MLLM model, and Wan-2.2-T2V-5B~\citep{wan2025wan} as the diffusion backbone. The training data are collected from diverse open-source resources, consisting of 10M videos and 20M images across various sources. 
The video pretraining data are collected from Sekai~\citep{li2025sekai}, FineVideo~\citep{Farré2024FineVideo}, Mira~\citep{ju2024miradata}, OpenVid~\citep{nan2024openvid}, InternVid~\citep{wang2023internvid}, Panda~\citep{chen2024panda}, and Koala~\citep{wang2025koala}. 
From this collection, we further filter a high-quality subset of 2M videos for downstream training. 
The image pretraining data are collected from LAION~\citep{schuhmann2022laion}. 
The EMA-based vision tokenizer is trained on the full image and video data. 
The MLLM is trained in three stages, including image pretraining, video pretraining, and final supervised fine-tuning (SFT) on the 2M high-quality video subset, which is also used for our ablation studies. The video DiT is trained on the same 2M subset, while the image DiT is trained on 5M images sampled from LAION. Additional implementation details are provided in Appendix~\ref{sec:implementation}.


\paragraph{Evaluation.}
We use the VBench-Long~\citep{huang2024vbench} benchmark to evaluate video generation performance, and the DAVIS~\citep{davis2019} dataset for video reconstruction performance.

\paragraph{Baselines.}
To ensure controlled comparisons, we begin with a DiT-only text-to-video baseline, Wan2.2-5B-FT, which is fine-tuned on the same training data as our model. We then investigate a range of intermediate representations, including refined prompts, implicit latent tokens, and explicit visual tokens. Based on these variants, we further evaluate different choices of visual representation spaces, MLLM-based generation mechanisms, and MLLM-DiT fusion strategies.

\subsection{Results on Different Intermediate Representations}

\begin{table*}[t!]
\caption{
Effectiveness of MLLM-DiT for Generation on VBench-Long Benchmark.}
\label{tab:main_vbench}
\centering
\resizebox{\textwidth}{!}{
\begin{tabular}{lccccc}
\toprule
\textbf{Method} 
& \textbf{MLLM Signal} 
& \textbf{Fusion Strategy} 
& \textbf{Quality$\uparrow$} 
& \textbf{Semantic$\uparrow$} 
& \textbf{Total Score$\uparrow$} \\
\midrule
DiT-only (Wan2.2-FT)
& Text 
& Text-condition
& 79.81 & 55.16 & 74.88 \\

\textit{+ Prompt Refinement} 
& Text 
& Text-condition
& 80.27 & 60.74 & 76.36 \\

\textit{+ Latent Bridging} 
& w/ Learnable Token
& Connector 
& 76.14 & 32.58 & 67.43 \\

\textit{+ Latent Bridging} 
& w/o Learnable Token
& Connector 
& 77.04 & 35.23 & 68.68 \\

\textit{+ Visual Tokens} 
& Discrete Token 
& Cross-attn
& \underline{81.74} & \underline{63.95} & \underline{78.18} \\

\rowcolor{gray!20}
\textit{+ Visual Tokens} 
& Discrete Token 
& Multi-cross-attn 
& \textbf{81.90} & \textbf{76.25} & \textbf{80.77} \\
\bottomrule
\end{tabular}
}
\end{table*}



Table~\ref{tab:main_vbench} compares different MLLM-DiT interfaces on VBench-Long. Prompt refinement improves the DiT-only baseline, mainly through better semantic fidelity and text alignment. This confirms that stronger text understanding is useful for video generation. Latent bridging performs worse than the DiT-only baseline. We attribute this degradation to the mismatch between the MLLM hidden states and the condition distribution expected by the DiT. Bridging this gap requires a connector that learns a complex mapping, which inherently demands substantial training data. The explicit discrete visual tokens generated by the MLLM achieve much stronger results, indicating that they provide useful structural guidance beyond text-only conditioning. When these tokens are injected into the first DiT block through cross-attention, the total score improves to 78.18. Furthermore, to strengthen the conditioning signal, we introduce multi-layer cross-attention, which injects visual tokens into multiple DiT blocks, further improving semantic performance and serving as our final design choice.

Beyond the controlled comparisons, we also report a contextual comparison with representative video generation systems in Appendix~\ref{sec:vbench}, Table~\ref{tab:vbench-long-t2v-new}. Although our method does not surpass these large-scale video generation systems, it achieves competitive performance on several dimensions, including subject, background, temporal, and appearance. These comparisons are provided for reference only, as the model scale and training data are not controlled across different systems.

\subsection{Ablation on the Space of Visual Tokens}

\begin{table*}[t!]
\centering
\caption{Evaluation of Codebook Intrinsic Properties.}
\resizebox{\textwidth}{!}{
\begin{tabular}
{p{3.7cm}p{2cm}p{2.7cm}p{2.9cm}p{2.9cm}p{2cm}}
\toprule
\textbf{Codebook} & \textbf{Entropy↑} & \textbf{Active Ratio↑} & \textbf{Top-10 Ratio↓} & \textbf{Quantize Error↓} \\
\midrule
Embedding & 0.321 & 0.052 & 0.704 & \underline{0.551} \\
Embedding + Decoder & 0.641 & 0.458 & 0.243 & 0.586 \\
Embedding + LLM & \underline{0.797} & \textbf{0.974} & \underline{0.186} & 0.938 \\
\midrule
\rowcolor{gray!20}
EMA & \textbf{0.812} & \underline{0.841} & \textbf{0.146} & \textbf{0.130} \\
\bottomrule
\end{tabular}
}
\label{tab:codebook_eval}
\end{table*}


To choose the best quantization methods, we evaluate intrinsic properties of the different codebooks.
As shown in Table~\ref{tab:codebook_eval}, the EMA-based tokenizer exhibits the lowest quantization error and the highest entropy, indicating both accurate reconstruction and balanced code utilization.
We further evaluate different codebooks in downstream MLLM prediction and DiT reconstruction tasks. Table~\ref{tab:vq-dit-recon-ablations} shows that the EMA-based codebook achieves the best reconstruction performance on the DAVIS dataset, consistent with its lower quantization error. Table~\ref{tab:vq-dit-gen-ablations} further demonstrates that it is also the most suitable for MLLM prediction, maintaining feature statistics closer to the original continuous visual representations and improving compatibility with the pretrained MLLM.


\subsection{Effect of the MLLM Generation Mechanism}

\begin{table*}[t!]
\caption{Ablation of MLLM-based Generation on VBench-Long Benchmark.}
\label{tab:ar-gen-ablations}
\centering
\resizebox{\textwidth}{!}{
\begin{tabular}{p{3cm}p{3cm}p{1.2cm}p{1.8cm}p{2cm}p{2.2cm}}
\toprule
\textbf{Input} & \textbf{Attention} & \textbf{Loss} & \textbf{Quality↑} & \textbf{Semantic↑} & \textbf{Total Score↑} \\
\midrule
Continuous Token & Causal Attn & MSE & 78.33 & 37.65 & 70.19 \\
Continuous Token& Bidirectional Attn & MSE & 77.24 & 36.77 & 69.14 \\
Discrete Token & Bidirectional Attn & CE & \textbf{79.63} & \underline{46.26} & \textbf{72.95} \\
\midrule
\rowcolor{gray!20}
Discrete Token & Causal Attn & CE & \underline{78.81} & \textbf{48.74} & \underline{72.79} \\
\bottomrule
\end{tabular}
}
\end{table*}

Table~\ref{tab:ar-gen-ablations} shows the results of causal attention and bidirectional attention for modeling visual tokens on VBench-Long. 
Compared with continuous inputs, discrete inputs consistently perform better under the same training budget. 
Moreover, using causal attention to model vision tokens provides stronger semantic following than bidirectional attention, with negligible impact on visual quality. Overall, these results suggest that discrete tokens with causal attention, modeled in the same autoregressive manner as text, can effectively produce high-quality intermediate visual representations for diffusion.

\subsection{Ablation on MLLM-DiT Fusion Strategy}


\begin{table*}[t!] 
\caption{Ablation of VQ and DiT for Reconstruction on DAVIS Dataset.}
\label{tab:vq-dit-recon-ablations}
\centering
\resizebox{\textwidth}{!}{
\begin{tabular}{p{3.8cm}p{3.8cm}p{2cm}p{2cm}p{2cm}}
\toprule
\textbf{Codebook} & \textbf{Injection Module} & \textbf{PSNR↑} & \textbf{SSIM↑} & \textbf{LPIPS↓}  \\
\midrule
Embedding             & Self Attention                 & 12.50 & 0.338 & 0.628  \\
Embedding + Decoder   & Self Attention               & 13.07 & 0.347 & 0.574  \\
Embedding + LLM       & Self Attention                & 11.94 & 0.290 & 0.715  \\
EMA     & Self Attention  & \underline{13.71} & \underline{0.358} & \underline{0.542} \\
\midrule
\rowcolor{gray!20}
EMA   & Cross Attention   & \textbf{14.05} & \textbf{0.388} & \textbf{0.528} \\
\bottomrule
\end{tabular}
}
\end{table*}

\begin{table*}[t!]
\caption{Ablation of VQ and DiT for Generation on VBench-Long Benchmark.}
\label{tab:vq-dit-gen-ablations}
\centering
\resizebox{\textwidth}{!}{
\begin{tabular}{p{3.8cm}p{3.3cm}p{2cm}p{2cm}p{2.5cm}}
\toprule
\textbf{Codebook} & \textbf{Injection Module} & \textbf{Quality↑} & \textbf{Semantic↑} & \textbf{Total Score↑} \\
\midrule
Embedding & Self Attention  & 79.07 & 45.80 & 72.41 \\
Embedding + Decoder & Self Attention & \textbf{79.20} & 40.82 & 71.52 \\
Embedding + LLM  & Self Attention & 75.72 & 34.68 & 67.51 \\
EMA  & Self Attention & \underline{78.81} & \underline{48.74} & \underline{72.79} \\
\midrule
\rowcolor{gray!20}
EMA  & Cross Attention  & 78.72 & \textbf{49.43} & \textbf{72.86} \\
\bottomrule
\end{tabular}
}
\vspace{-1em}
\end{table*}

We further evaluate how the DiT should fuse the explicit visual tokens. As shown in Table~\ref{tab:vq-dit-recon-ablations} and Table~\ref{tab:vq-dit-gen-ablations}, discrete tokens achieve strong reconstruction performance and semantic generation capability. Adding a decoder reconstruction loss yields slightly higher overall generation quality but incurs lower semantic scores. In addition, single-layer self-attention and cross-attention yield comparable performance in both reconstruction and generation, suggesting that both fusion mechanisms can effectively leverage MLLM-produced visual tokens. More ablation studies on different model scales and classifier-free guidance (CFG)~\citep{ho2022classifier} are provided in Appendix~\ref{sec:more_ablations}.

\begin{table*}[t!]
\caption{Comparison of Latency and Memory Usage.}
\label{tab:latency}
\centering
\resizebox{\textwidth}{!}{
\begin{tabular}{lcccc}
\toprule
\textbf{Model} & \textbf{Denoising Steps} & \textbf{Latency (s) ↓} & \textbf{Memory (GB) ↓} & \textbf{Total Score on VBench-Long ↑} \\
\midrule
DiT-only & 50 & 37 & 29 (Peak) & 74.88 \\
MLLM + DiT & 50 & 45 & 29 (Peak) & 78.18 \\
MLLM + DiT & 30 & 31 & 29 (Peak) & 77.27 \\
\bottomrule
\end{tabular}
}
\end{table*}

\begin{figure}[!t]
  \centering
  \includegraphics[width=1.0\linewidth]{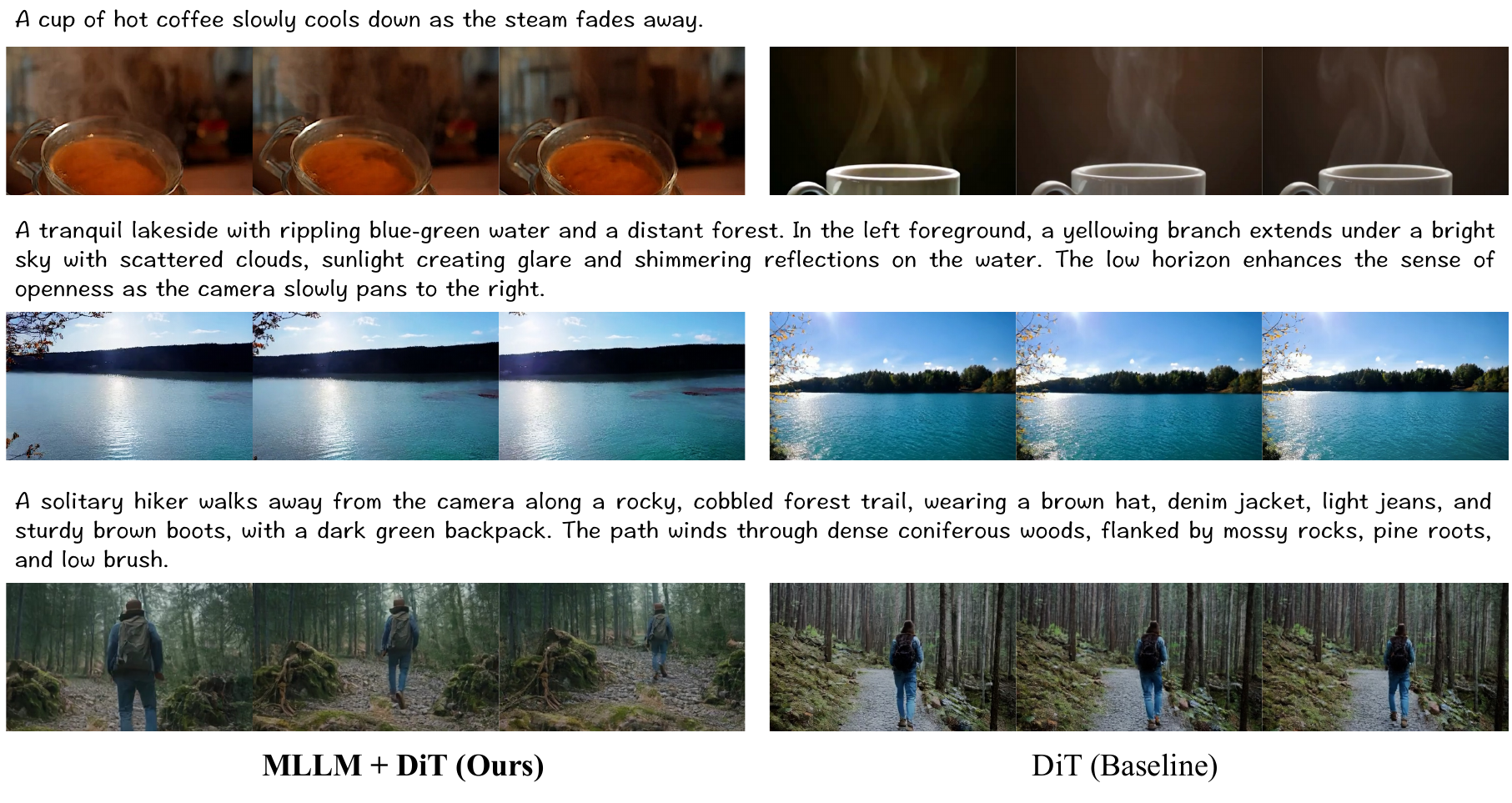}
  \vspace{-18pt}
  \caption{\textbf{Qualitative Comparisons.} MLLM+DiT yields better causal transitions, temporal consistency, and semantic alignment than the DiT-only baseline in these examples.}
  \label{fig:compare}
\end{figure}

\subsection{Qualitative Analysis}

We present a qualitative comparison in Figure~\ref{fig:compare}. MLLM+DiT better captures causal transitions described in the prompts and reflects them in the generated videos, e.g., the steam fades away and the camera slowly pans to the right. In addition, it demonstrates strong text understanding by accurately rendering fine-grained details from the prompts, such as a brown hat, a dark green backpack, and mossy rocks.
We additionally present representative image generation results in Figure~\ref{fig:z_image_generation}, showing high-quality images across diverse scenes, ranging from natural landscapes to indoor environments and objects. Our method accurately captures both global scene composition and fine-grained details, including accurate text rendering. Additional qualitative comparisons are provided in Appendix~\ref{sec:more_vis}.

\subsection{Efficiency Analysis}

The MLLM brings an additional semantic planning step before diffusion rendering. As shown in Table~\ref{tab:latency}, MLLM+DiT increases inference latency from 37s to 45s compared with the DiT-only baseline at 50 denoising steps, while maintaining the same peak memory usage. Reducing denoising steps from 50 to 30 decreases latency to 31s, with the VBench-Long total score dropping only from 78.18 to 77.27, indicating a favorable quality-latency trade-off under a reduced sampling budget.

\section{Conclusion}



In this work, we systematically study how an MLLM should be integrated with a DiT for video generation. Through three key questions covering intermediate representation, generation mechanism, and fusion strategy, we identify that discrete semantic tokens from an EMA-based tokenizer, causal autoregressive generation, and multi-layer cross-attention conditioning form an effective design. Finally, we propose BiVidGen, which leverages explicit MLLM semantic planning to guide DiT-based rendering, achieving clear improvements over the DiT-only baseline on VBench-Long. We believe our findings offer useful insights for future research on hybrid video generation architectures.

\bibliographystyle{plainnat}  
\bibliography{neurips_2026}

\newpage
\appendix

\renewcommand{\appendixpagename}{Appendix}
\appendixpage
\startcontents[sections]
\section*{Table of Contents}
\vspace{-15pt}\noindent\hrulefill

\printcontents[sections]{l}{1}{\setcounter{tocdepth}{2}}

\noindent\hrulefill
\newpage

\section{Additional Implementation Details}
\label{sec:implementation}

\subsection{Vision Tokenizer}
We use a codebook of size 16,384 with a code dimension of 2,048. The EMA update is applied with a decay constant of $\mu = 0.99$ to ensure stable and smooth codebook updates. To prevent codebook collapse and improve utilization, unused codes are periodically reset every 20 training steps.

\subsection{MLLM-Based Semantic Generator}
We use Qwen3-VL-2B~\citep{bai2025qwen3vltechnicalreport} as the base model. Image pretraining uses a batch size of 256 with a learning rate decayed from $2 \times 10^{-5}$ to $5 \times 10^{-6}$ via cosine scheduling and a 0.03 warm-up ratio. Video pretraining uses a batch size of 64 with a constant learning rate of $5 \times 10^{-6}$. Continued training on the 2M high-quality video subset uses a batch size of 64 and a learning rate decayed from $5 \times 10^{-6}$ to $1 \times 10^{-7}$ with cosine scheduling. Weight decay for the three stages is 0.01, 0.01, and 0.0, respectively. Training data are resized and cropped to $288 \times 512$, $384 \times 384$, and $512 \times 288$, with 1 frame for images and 121 frames for videos. All training stages use the AdamW optimizer~\citep{loshchilov2017decoupled} with $\beta_1 = 0.9$ and $\beta_2 = 0.99$. During inference, the sampling temperature is set to 0.98 to balance visual quality and motion dynamics.

\subsection{DiT-Based Pixel Renderer}
We adopt Wan-2.2-5B~\citep{wan2025wan} and Z-Image-6B~\citep{cai2025z-image} as diffusion-based renderers for video and image generation, respectively. Both models are trained with a learning rate of $2 \times 10^{-5}$, a warm-up ratio of 0.03, and a weight decay of 0.03. All experiments are conducted on 32 NVIDIA A100-40G GPUs using DeepSpeed ZeRO-2~\citep{rajbhandari2020zero} to reduce memory consumption. During training, we apply drop rates of 0.1 and 0.2 for text and visual (from our EMA-based tokenizer) conditions, respectively, to improve model robustness and prevent over-reliance on specific modalities. At inference, we use 50 denoising steps and apply classifier-free guidance (CFG)~\citep{ho2022classifier} with scales of 1.5 for both text and visual conditions, which provides a balanced trade-off between controllability and visual fidelity in the generated results.

\section{Image Generation Results}
\label{sec:image_captions}

\begin{figure}[!t]
  \centering
  \includegraphics[width=1.0\linewidth]{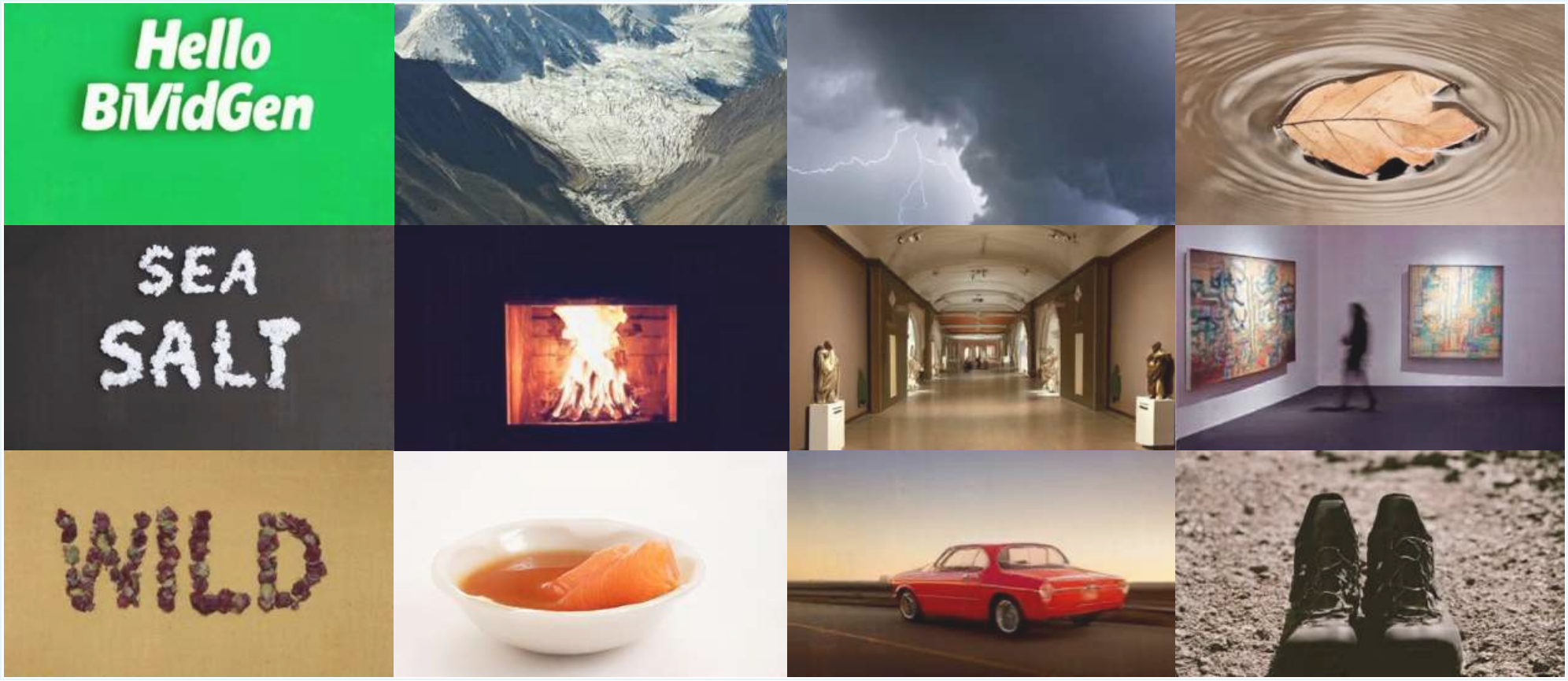}
  \vspace{-12pt}
\caption{\textbf{Image Generation Results.} Our method generates visually coherent images across diverse scenes with fine-grained details, including accurate text content. Prompts are provided in Table~\ref{tab:image_prompts}.} 
  \label{fig:z_image_generation}
\end{figure}

We present representative image generation results in Figure~\ref{fig:z_image_generation}. Our method can generate high-quality images across various scenarios, accurately capturing fine-grained details, including accurate text rendering (e.g., ``BiVidGen").
Table~\ref{tab:image_prompts} provides the detailed descriptive captions for the generated images in Figure~\ref{fig:z_image_generation}, organized according to their corresponding row and column positions.

\section{Details of Hierarchical Token Structures}
\label{sec:hierarchical_token}

The main motivation for exploring hierarchical token structures stems from the base model, Qwen3-VL-2B~\citep{bai2025qwen3vltechnicalreport}, which leverages hierarchical features from the vision transformer (ViT)~\citep{dosovitskiy2020vit} to enhance visual understanding. These hierarchical features have been shown in their technical report to consistently improve visual understanding performance across several related benchmarks. Naturally, we aim to investigate whether incorporating features from shallow ViT layers can similarly benefit video generation performance in a controllable manner. Our pipeline involves training hierarchical discrete EMA-based tokenizers, using the MLLM to generate hierarchical discrete visual tokens, and finally conditioning the DiT renderer on these hierarchical features. We describe the detailed design and implementation in the following.

\begin{figure*}[!t]
  \centering
  \begin{minipage}[t]{0.5\textwidth}
    \centering
    \includegraphics[width=\linewidth]{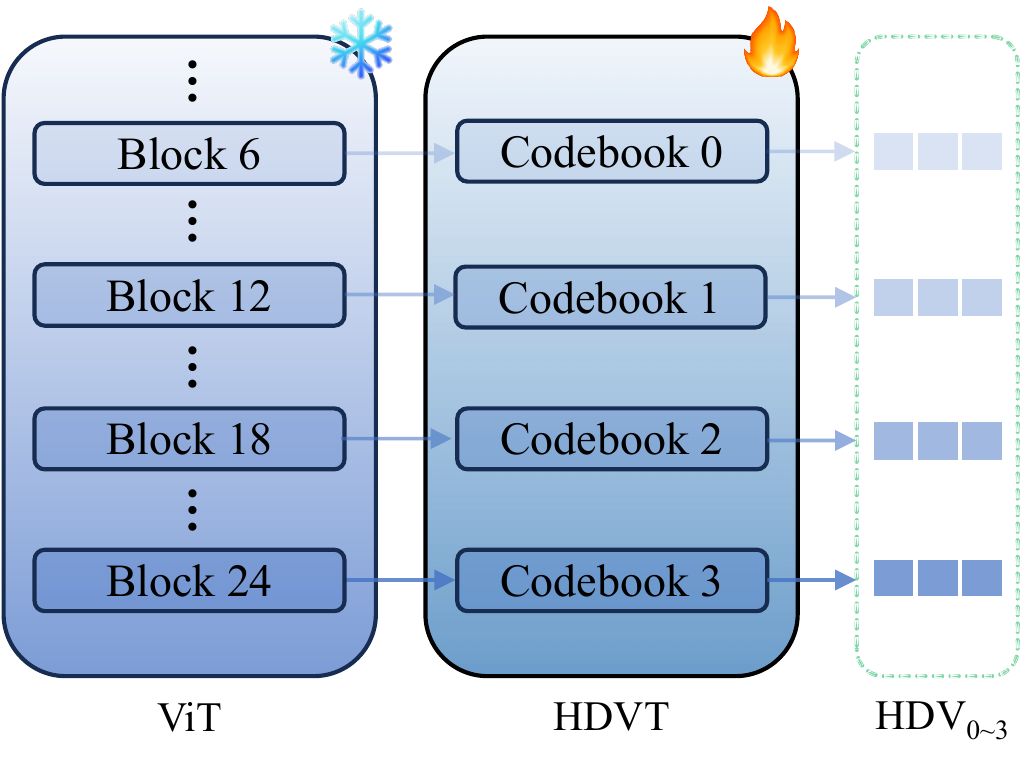}
    \caption{\textbf{Architecture of Hierarchical EMA-Based Visual Tokenizer.} We quantize features from different layers of the ViT to capture semantic information at different levels of granularity.}
    \label{fig:vq}
  \end{minipage}
  \hfill
  \begin{minipage}[t]{0.45\textwidth}
    \centering
    \includegraphics[width=\linewidth]{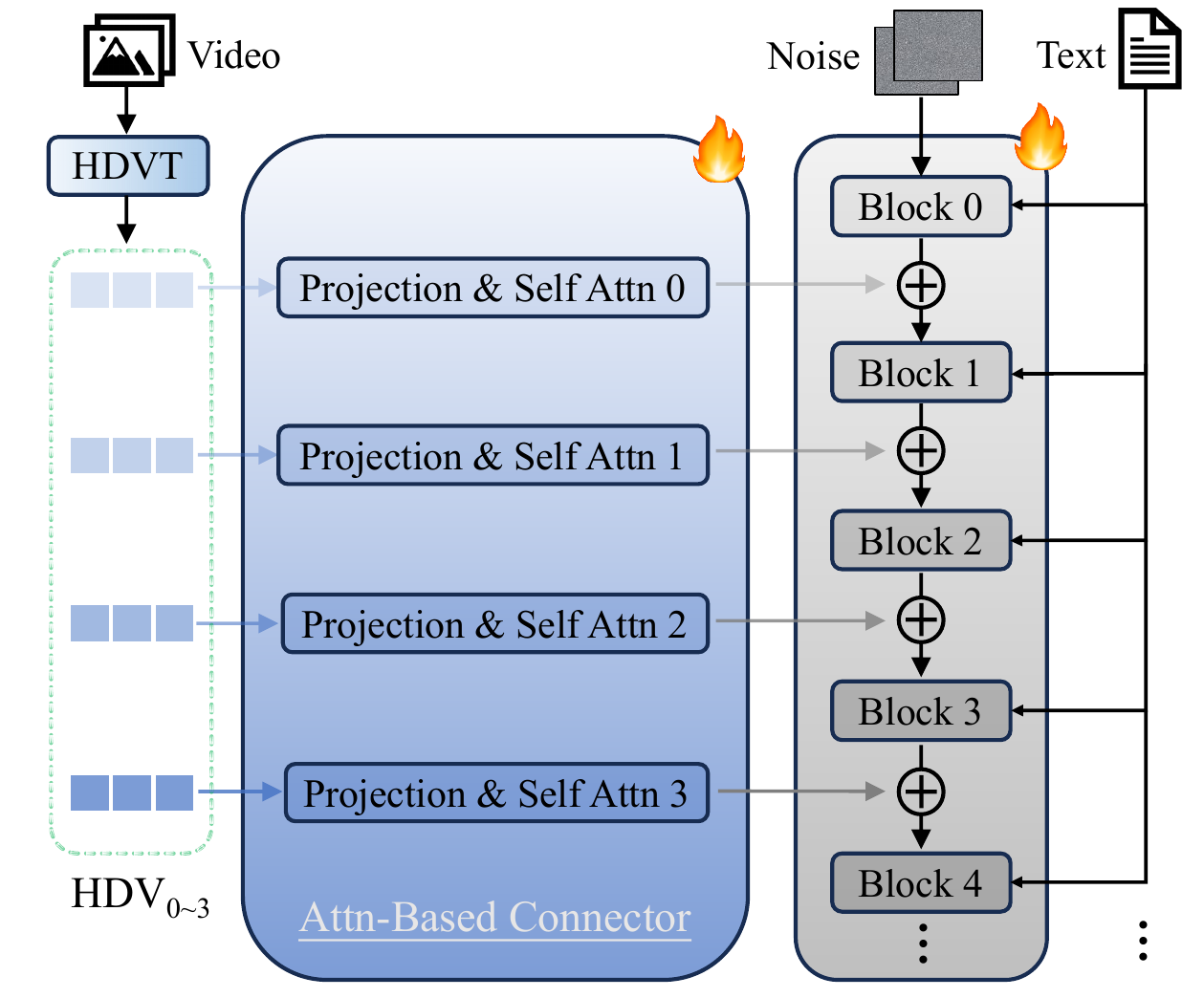}
    \caption{\textbf{DiT Reconstruction Based on Hierarchical Token Structures.} The DiT leverages semantic information at different levels of granularity for reconstruction.}
    \label{fig:dit}
  \end{minipage}
\end{figure*}

First, we train the hierarchical tokenizer (Figure~\ref{fig:vq}). Instead of quantizing only the last-layer feature of the ViT, we quantize multiple intermediate layer features evenly to produce hierarchical discrete visual representations, ranging from low-level structures to high-level semantics. This process uses independent codebooks for different ViT layers to better preserve feature diversity. Next, we employ the MLLM to generate discrete hierarchical visual tokens, implemented with separate generation heads for each token level. Finally, these hierarchical features are fed into different layers of the DiT to serve as reconstruction conditions (Figure~\ref{fig:dit}). We also explore various injection strategies, such as injecting all features into the first block, sequentially across blocks (e.g., 0, 1, 2, 3), or evenly spaced across blocks (e.g., 0, 8, 16, 24). Our experiments in Table~\ref{tab:single-multiple} indicate that sequential injection yields the most stable and overall best results for both reconstruction and generation. 

Compared to single-level visual tokens (using only the last ViT layer), hierarchical token structures slightly improve reconstruction performance on the DAVIS~\citep{davis2019} dataset, as shallow-layer features provide additional fine-grained structural details that are beneficial for reconstruction tasks. However, they result in worse generation performance on VBench-Long~\citep{huang2024vbench} compared to single-level tokens. This is likely because hierarchical visual tokens significantly increase the prediction difficulty for the MLLM, making training less stable. Errors in predicting shallow-layer tokens can propagate and negatively affect deeper-layer token prediction, thereby reducing overall generation quality. Consequently, we do not adopt this architecture in our final model, although hierarchical tokens remain an interesting and promising direction for future exploration. They inevitably introduce additional computational cost and architectural complexity, and may require more careful modeling or more advanced designs to fully realize their potential.

\begin{table*}[t!]
\caption{Comparison With Hierarchical Discrete Tokens.}
\label{tab:single-multiple}
\centering
\resizebox{\textwidth}{!}{
\begin{tabular}{p{2.3cm}p{3.5cm}p{2cm}p{1.8cm}p{2cm}p{2.2cm}}
\toprule
\textbf{Benchmark} & \textbf{Input} & \textbf{Layers} & \textbf{Quality↑} & \textbf{Semantic↑} & \textbf{Total Score↑} \\
\midrule
\multirow{4}{*}{VBench-Long} &
Discrete & 0 & 78.81 & 48.74 & 72.79 \\
& Discrete Hierarchical & 0, 0, 0, 0 & 79.68 & 42.93 & 72.33 \\
& Discrete Hierarchical & 0, 8, 16, 24 & 79.04 & 41.45 & 71.52 \\
& Discrete Hierarchical & 0, 1, 2, 3 & 79.48 & 44.78 & 72.54 \\
\bottomrule
\end{tabular}
}
\resizebox{\textwidth}{!}{
\begin{tabular}{p{2cm}p{3.5cm}p{2cm}p{1.5cm}p{1.5cm}p{1.5cm}}
\toprule
\textbf{Benchmark} & \textbf{Input} & \textbf{Layers} & \textbf{PSNR↑} & \textbf{SSIM↑} & \textbf{LPIPS↓} \\
\midrule
\multirow{5}{*}{DAVIS} &
Continuous & 0 & 17.49 & 0.467 & 0.399 \\
& Discrete & 0 & 13.71 & 0.358 & 0.542 \\
& Discrete Hierarchical  & 0, 0, 0, 0   & 15.84 & 0.430 & 0.455 \\
& Discrete Hierarchical  & 0, 8, 16, 24  & 13.39 & 0.376 & 0.572 \\
& Discrete Hierarchical  & 0, 1, 2, 3    & 16.02 & 0.436 & 0.426 \\
\bottomrule
\end{tabular}
}
\end{table*}

\begin{table}[t!]
\caption{Detailed Prompts for the Images in Figure~\ref{fig:z_image_generation}.}
\centering
\small
\begin{tabular}{p{1cm}p{1cm}p{11cm}}
\hline
\textbf{Row} & \textbf{Col} & \textbf{Caption} \\ \hline
1 & 1 & A bright green background featuring the white text "Hello BiVidGen". \\
1 & 2 & A sweeping aerial landscape of a massive valley glacier. \\
1 & 3 & A dramatic photo of a lightning strike illuminating dark storm clouds. \\
1 & 4 & A dried brown leaf floating on water, creating concentric ripples. \\ \hline
2 & 1 & A photo of the words 'SEA SALT' formed by coarse white salt crystals on a dark slate board. \\
2 & 2 & A warm, bright fire burning within a dark brick fireplace. \\
2 & 3 & A long, symmetrical view of a classical museum hall with statues. \\
2 & 4 & A blurred silhouette of a person walking past modern paintings in a gallery. \\ \hline
3 & 1 & The words 'WILD' spelled out using real, colorful pressed petals on handmade paper. \\
3 & 2 & Fresh slices of raw salmon in a small white bowl of soy sauce. \\
3 & 3 & A vintage red sports car driving rapidly along a coastal highway at sunset. \\
3 & 4 & A pair of sturdy, worn hiking boots resting on rocky ground. \\ \hline
\end{tabular}
\label{tab:image_prompts}
\end{table}

\section{Contextual Comparison With Representative Methods}
\label{sec:vbench}

As shown in Table~\ref{tab:vbench-long-t2v-new}, our method achieves competitive performance on the VBench-Long benchmark~\citep{huang2024vbench}, with strong results on several key dimensions, including background consistency and appearance style. These results indicate that incorporating the MLLM as a semantic planner effectively helps maintain stable video structures and improves semantic alignment with text prompts, leading to clear and consistent improvements over the DiT-only baseline (Table~\ref{tab:main_vbench}).

We observe relatively lower scores on some metrics (e.g., aesthetic quality, human action), which can be attributed to multiple factors. First, the limited quantity and quality of our training data, sourced from open resources, constrain overall performance. Second, the capacity and performance of the base model also play an important role. In addition, due to constrained training resources, we adopt relatively low training resolutions, which further affects the baseline performance. While integrating the MLLM significantly improves semantic alignment, it remains challenging to fully compensate for these limitations, preventing the model from completely reaching the current state-of-the-art.

\begin{figure*}[!t]
  \centering
  \includegraphics[width=0.99\linewidth]{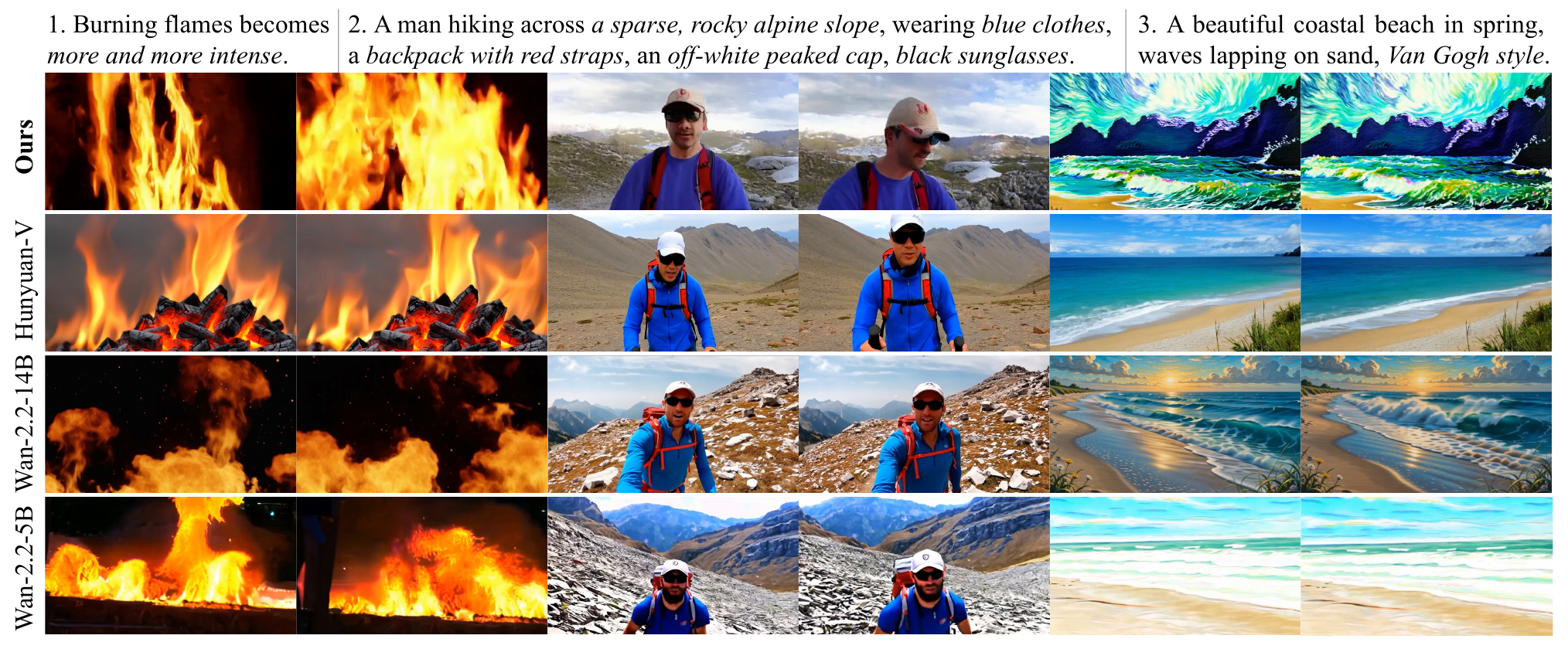}
   \caption{\textbf{Video Comparisons With Existing Methods.} With the help of MLLM, our BiVidGen demonstrates clear advantages in prompt adherence and superior video quality.}
   \label{fig:compare_sota}
\end{figure*}

\begin{table*}[t!]
\caption{T2V Performance on VBench-Long~\citep{huang2024vbench} Benchmark. \textbf{Bold} indicates the best results, \underline{underline} indicates the second best, and \textit{italic} indicates the third best.}
\label{tab:vbench-long-t2v-new}
\centering
\resizebox{\textwidth}{!}{%
\begin{tabular}{lcccccccccc}
\toprule
\multirow{2}{*}{\textbf{Method}} & \multicolumn{3}{c}{\textbf{Overall Scores}} & \multicolumn{5}{c}{\textbf{Technical Quality}} & \multicolumn{2}{c}{\textbf{Aesthetic Quality}} \\
\cmidrule(lr){2-4} \cmidrule(lr){5-9} \cmidrule(lr){10-11}
 & Total Score↑ & Quality↑ & Semantic↑ & Subject↑ & Background↑ & Temporal↑ & Motion↑ & Dynamic↑ & Aesthetic↑ & Imaging↑ \\
\midrule
Sora \citep{liu2024sora} & 84.28 & 85.51 & 79.35 & 96.23 & 96.35 & 98.87 & 98.74 & 79.91 & 63.46 & 68.28 \\ 
OpenSora \citep{zheng2024opensora} & 79.76 & 81.35 & 73.39 & 96.75 & 97.61 & 99.53 & 98.50 & 42.39 & 56.85 & 63.34 \\
HunyuanVideo \citep{kong2024hunyuanvideo} & 83.24 & 85.86 & 75.82 & 97.32 & 97.93 & 99.49 & 98.99 & 70.83 & 60.36 & 67.56 \\
Veo-3 \citep{wiedemer2025veo} & 85.06 & 85.70 & 82.49 & 97.36 & 96.89 & 99.30 & 99.16 & 72.43 & 63.81 & 68.23 \\
Gen-3 \citep{gen2024} & 82.32 & 84.11 & 75.17 & 97.01 & 96.62 & 99.61 & 99.23 & 60.14 & 63.34 & 66.82 \\
Wan2.2-T2V-14B \citep{wan2025wan} & 82.61 & 85.03 & 72.92 & 97.46 & 96.67 & 98.92 & 97.93 & 68.24 & 63.52 & 71.69 \\
\midrule
\rowcolor{gray!20}
\textbf{Ours} & 80.77 & 81.90 & \textit{76.25} & \textbf{97.74} & \textbf{98.02} & \textbf{99.68} & 98.06 & 46.94 & 57.14 & 62.70 \\
\bottomrule
\end{tabular}%
}
\resizebox{\textwidth}{!}{%
\begin{tabular}{lccccccccc}
\toprule
\multirow{2}{*}{\textbf{Method}} & \multicolumn{9}{c}{\textbf{Semantic Fidelity}} \\
\cmidrule(lr){2-10}
 & Object↑ & Multi-Obj↑ & Action↑ & Color↑ & Spatial↑ & Scene↑ & Appearance↑ & Temporal↑ & Overall↑ \\
\midrule
Sora \citep{liu2024sora} & 93.93 & 70.85 & 98.20 & 80.11 & 74.29 & 56.95 & 24.76 & 25.01 & 26.26 \\
OpenSora \citep{zheng2024opensora} & 82.22 & 51.83 & 91.20 & 90.08 & 68.56 & 42.44 & 23.95 & 24.54 & 26.85 \\
HunyuanVideo \citep{kong2024hunyuanvideo} & 86.10 & 71.66 & 93.42 & 91.60 & 68.09 & 53.69 & 19.80 & 23.89 & 26.44 \\
Veo-3 \citep{wiedemer2025veo} & 93.89 & 82.20 & 99.40 & 82.48 & 84.26 & 57.43 & 23.55 & 25.97 & 27.88 \\
Gen-3 \citep{gen2024} & 87.81 & 53.64 & 96.40 & 80.90 & 65.03 & 54.57 & 24.31 & 24.71 & 26.69 \\
Wan2.2-T2V-14B \citep{wan2025wan} & 85.04 & 74.61 & 83.40 & 88.93 & 80.22 & 34.38 & 20.27 & 23.13 & 24.68 \\
\midrule
\rowcolor{gray!20}
\textbf{Ours} & 83.07 & 57.27 & 89.25 & 85.28 & \textit{75.31} & \textit{54.60} & \textbf{25.12} & \underline{25.48} & 26.10 \\
\bottomrule
\end{tabular}%
}
\end{table*}

\section{Additional Ablations}
\label{sec:more_ablations}

We provide additional ablation studies to further analyze our design choices. First, we evaluate the scalability of our method across different model scales and consistently observe similar performance gains, demonstrating that our method generalizes beyond a specific model size. Second, we analyze the effects of classifier-free guidance scale and the number of denoising steps on generation quality.

\subsection{Scalability Across Model Sizes}

Due to resource constraints, training on a significantly larger backbone is infeasible. To assess scalability, we conduct experiments on Wan2.1-T2V-1.3B~\citep{wan2025wan} and consistently observe performance gains from MLLM-based semantic planning (Table~\ref{tab:more_ablation}). Notably, the improvement holds across two model scales (1.3B \& 5B), suggesting that the gains are not tied to a specific model size but instead complement the underlying backbone capacity. This indicates that our method is generally applicable and can be effectively integrated with models of different scales.

\subsection{CFG Scales and Denoising Steps}
We additionally study the impact of CFG scales and the number of denoising steps on video reconstruction performance. As shown in Table~\ref{tab:cfg_ablation}, a CFG scale of 1.5 provides the best trade-off between visual fidelity and controllability across different settings. Benefiting from the strong and informative visual tokens produced by the MLLM planner, our DiT can generate high-fidelity videos with a relatively small number of denoising steps (e.g., 10), showing only minor degradation compared to using 50 steps. This demonstrates that our method can improve sampling efficiency and potentially leads to reduced inference latency compared to conventional DiT-only generation.




\section{More Visualizations}
\label{sec:more_vis}

\begin{figure*}[!t]
  \begin{center}
    \centerline{\includegraphics[width=1\textwidth]{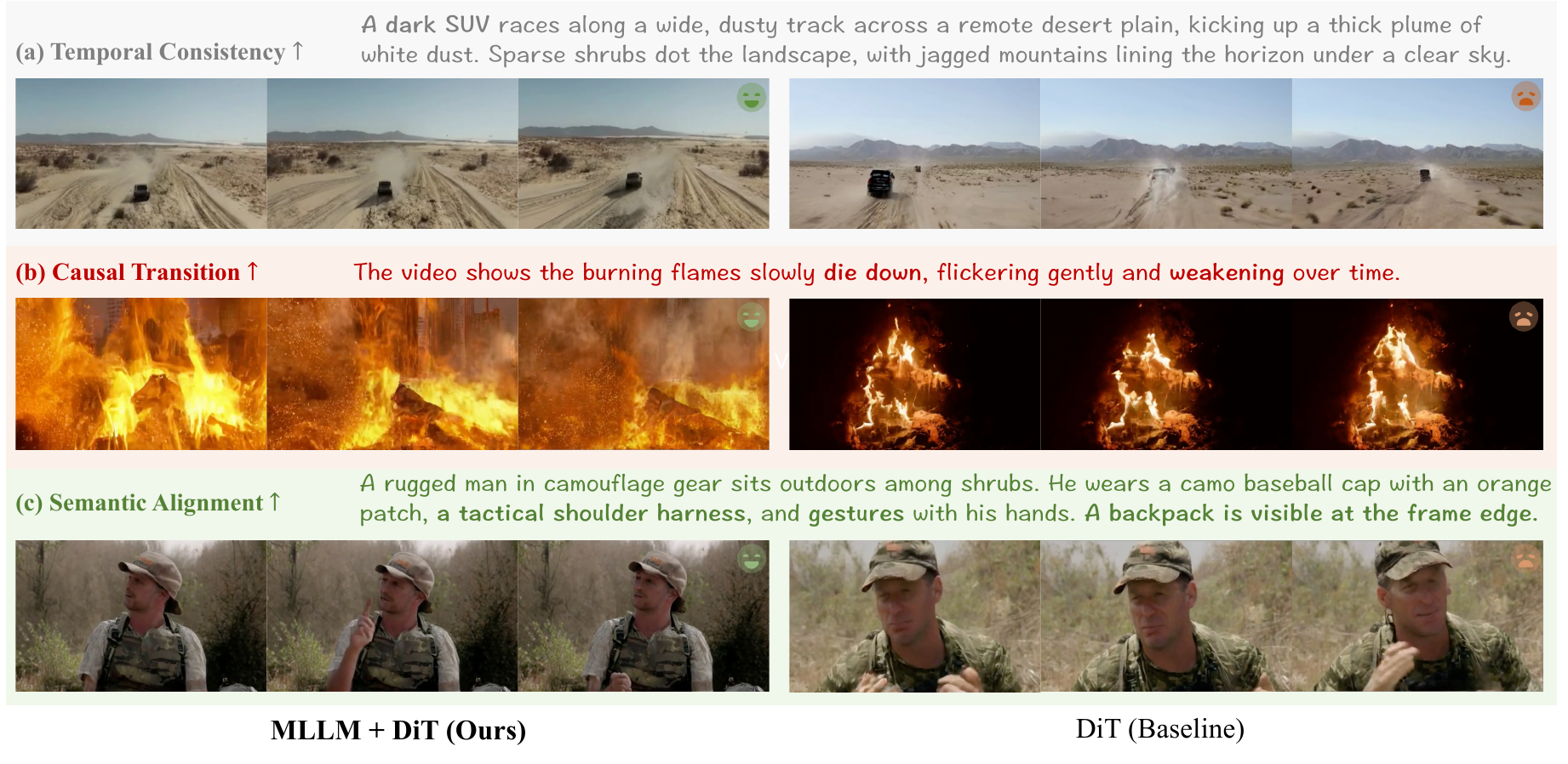}}
    \vspace{-5pt}
    \caption{
    \textbf{Video Comparisons With the Baseline.} Our method integrates an MLLM with a DiT for video generation. The semantic planning of the MLLM enables the DiT to generate better videos, especially in terms of temporal consistency, causal transition, and semantic alignment.
    }
    \label{fig:teaser}
  \end{center}
\end{figure*}

As shown in Figure~\ref{fig:compare_sota}, our method demonstrates clear advantages in prompt adherence and superior video quality compared with existing methods. Figure~\ref{fig:teaser} further shows that semantic planning of the MLLM enables the DiT to generate better videos,
especially in terms of temporal consistency, causal transition, and semantic alignment, compared with DiT-only generation.

We provide additional visualization results in the submitted zip file for further qualitative analysis. The archive contains two folders, \texttt{videos} and \texttt{images}, presenting the corresponding video and image cases. These examples cover a wide range of scenes and content types, further demonstrating the superior generation capability of BiVidGen across multiple resolutions and visual scenarios.

\section{Limitations}
\label{sec:limitation}

We briefly discuss the current limitations. Due to limited training resources and the availability of high-quality data, it remains challenging to achieve state-of-the-art performance on VBench-Long. In addition, our training resolutions are relatively low, restricted to 288×512, 384×384, and 512×288, which may further limit the final generation quality and fine-grained visual details.

\begin{table*}[t!]
\centering
\begin{minipage}{0.55\textwidth}
\centering
\caption{Scalability Test on VBench-Long.}
\label{tab:more_ablation}
\resizebox{\textwidth}{!}{
\begin{tabular}{lccc}
\toprule
\textbf{Model} & \textbf{Quality↑} & \textbf{Semantic↑} & \textbf{Total↑} \\
\midrule
Wan2.2-5B-FT & 79.81 & 55.16 & 74.88 \\
\textit{+ Prompt-Refinement} & 80.27 & 60.74 & 76.36 \\
\rowcolor{gray!20}
\textit{+ MLLM-Planning} & \textbf{81.74} & \textbf{63.95} & \textbf{78.18} \\
\midrule
Wan2.1-1.3B-FT & 79.06 & 52.49 & 73.75 \\
\textit{+ Prompt-Refinement} & 79.86 & 57.97 & 75.48 \\
\rowcolor{gray!20}
\textit{+ MLLM-Planning} & \underline{79.93} & \underline{60.88} & \underline{76.12} \\
\bottomrule
\end{tabular}
}
\end{minipage}
\hfill
\begin{minipage}{0.42\textwidth}
\centering
\caption{CFG Scale and Denoising Steps.}
\label{tab:cfg_ablation}
\resizebox{\textwidth}{!}{
\begin{tabular}{cccccc}
\toprule
\textbf{CFG} & \textbf{Steps} & \textbf{PSNR$\uparrow$} & \textbf{SSIM$\uparrow$} & \textbf{LPIPS$\downarrow$} \\
\midrule
1.0 & 50 & \textbf{13.73} & {0.352} & 0.564 \\
1.5 & 50 & \underline{13.71} & \textbf{0.358} & \underline{0.542} \\
1.5 & 30 & 13.47 & \underline{0.353} & 0.545 \\
1.5 & 10 & 12.54 & 0.334 & 0.561 \\
2.0 & 50 & 13.18 & 0.346 & \textbf{0.538} \\
2.5 & 50 & 12.90 & 0.342 & 0.540 \\
3.0 & 50 & 12.68 & 0.338 & 0.543 \\
\bottomrule
\end{tabular}
}
\end{minipage}
\end{table*}

\section{Broader Impacts}
\label{sec:broader_impacts}
This work improves text-to-video generation by using MLLMs for explicit semantic planning before DiT-based rendering. It may benefit visual generation by improving semantic alignment and temporal coherence. However, more controllable video generation may also increase risks such as misinformation, biased content, and copyright misuse. Responsible deployment should include provenance tracking, misuse detection, dataset auditing, and clear disclosure of synthetic content.

\section{Future Work}
\label{sec:future_work}

We plan to explore more advanced reasoning over semantic visual features to better leverage the capabilities of the MLLM, as well as to investigate more efficient and scalable architectures for both the MLLM-based semantic planner and the DiT renderer. We also consider extending our framework to higher-resolution generation and more diverse data sources to further improve performance.

\end{document}